\documentclass[uspaper, 10 pt, conference]{format/ieeeconf}

\IEEEoverridecommandlockouts                              

\usepackage{graphicx}
\usepackage[cmex10]{amsmath}
\usepackage{amssymb}
\usepackage{mathtools}
\usepackage{array}
\usepackage{booktabs}
\usepackage{url}
\usepackage{dblfloatfix}
\usepackage{cite}
\usepackage{csquotes}
\usepackage{tikz}

\newcommand{\shrteq}{\mathrel{\scalebox{0.73}[0.9]{$=$}}}

\title{\LARGE \bf
Towards Miniature Humanoid Tele-Loco-Manipulation using Virtual Reality and Reinforcement Learning
}

\author{Nicolas Kosanovic$^{1^*}$, Jordan Dowdy$^{1}$, and Jean Chagas Vaz$^{2}$
\thanks{$^{1^*}$N. Kosanovic and $^{1}$J. Dowdy are PhD students with the Department of Electrical and Computer Engineering, University of Louisville. \break 
$^*$direct all correspondence to this author. \break
{\tt\small nicolas.kosanovic@louisville.edu}, {\tt\small jordan.dowdy@louisville.edu}}
\thanks{$^{2}$Dr. Jean Chagas Vaz is with the Faculty of Electrical and Computer Engineering, University of Louisville, Louisville, KY, 40208, USA. {\tt\small jean.chagasvaz@louisville.edu}}
} 

\newcommand\copyrighttext{%
  \scriptsize\centering
  \textcopyright\ 2025 IEEE. Personal use of this material is permitted.
  Permission from IEEE must be obtained for all other uses, in any current or future
  media, including reprinting/republishing this material for advertising or promotional
  purposes, creating new collective works, for resale or redistribution to servers or
  lists, or reuse of any copyrighted component of this work in other works.\\
  Accepted manuscript. Published in the 2025 IEEE-RAS 24th International Conference on
  Humanoid Robots (Humanoids), pp. 1233--1240. doi:
  10.1109/Humanoids65713.2025.11264861.}
\newcommand\copyrightnotice{%
\begin{tikzpicture}[remember picture,overlay]
\node[anchor=south,yshift=10pt] at (current page.south) {\fbox{\parbox{\dimexpr\textwidth-\fboxsep-\fboxrule\relax}{\copyrighttext}}};
\end{tikzpicture}%
}

\begin{document}

\maketitle

\copyrightnotice

\thispagestyle{empty}
\pagestyle{empty}

\begin{abstract}
Full-sized humanoid robot capabilities have grown exponentially in recent years, aiming towards general-purpose deployment in human environments. A popular control method used by manufacturers utilizes Virtual Reality for upper-body teleoperation and Reinforcement Learning for lower-body balance and locomotion control. As a result, a single remote operator can see, manipulate, and navigate about a real, distant physical environment. This powerful control stack is often relegated to expensive full-sized robots, many of which are inaccessible to the research community. Miniature humanoids are more prevalent, but employ less biomimicry in their design (e.g. fewer sensors, Degrees of Freedom, etc) and lack similar developments. This paper describes a compliant full-body telepresence control stack developed \textit{from the ground up} for miniature humanoids. Framework experimentation on ROBOTIS OP3 hardware showcases walking at speeds up to $0.45$ $m/s$ independent of arm motions. Tele-loco-manipulation is demonstrated via a cube relocation experiment with an expert human operator. On average, the teleoperated system moved $2$ different $40$ $g$ cubes within $10$ $mins$, walking a total distance of $5$ $m$. Overall, the developed system shows potential for miniature humanoid tele-loco-manipulation.

\end{abstract}

\section{INTRODUCTION}
\hyphenpenalty=10000
Reinforcement Learning (RL) locomotion and Virtual Reality (VR) teleoperation have greatly expanded humanoid robot capabilities in recent years. This is especially evident within the commercial humanoid sector, with robot manufacturers using such methods for real-world deployment \cite{1X, Unitree}. Data generated from such employments is used within \textit{imitation learning} schemes, enabling machines to accomplish tasks purely from high-level instructions \cite{nuclearSuperPaper}. Such technology could transform how physical labor is done, potentially eliminating the need for humans within \textit{Dull, Dirty, and Dangerous} work like firefighting, high-risk manufacturing, or unexploded ordinance defusal.

\begin{figure}[t]
	\centering
	\vspace{2mm}
	\includegraphics[width= 3.4in]{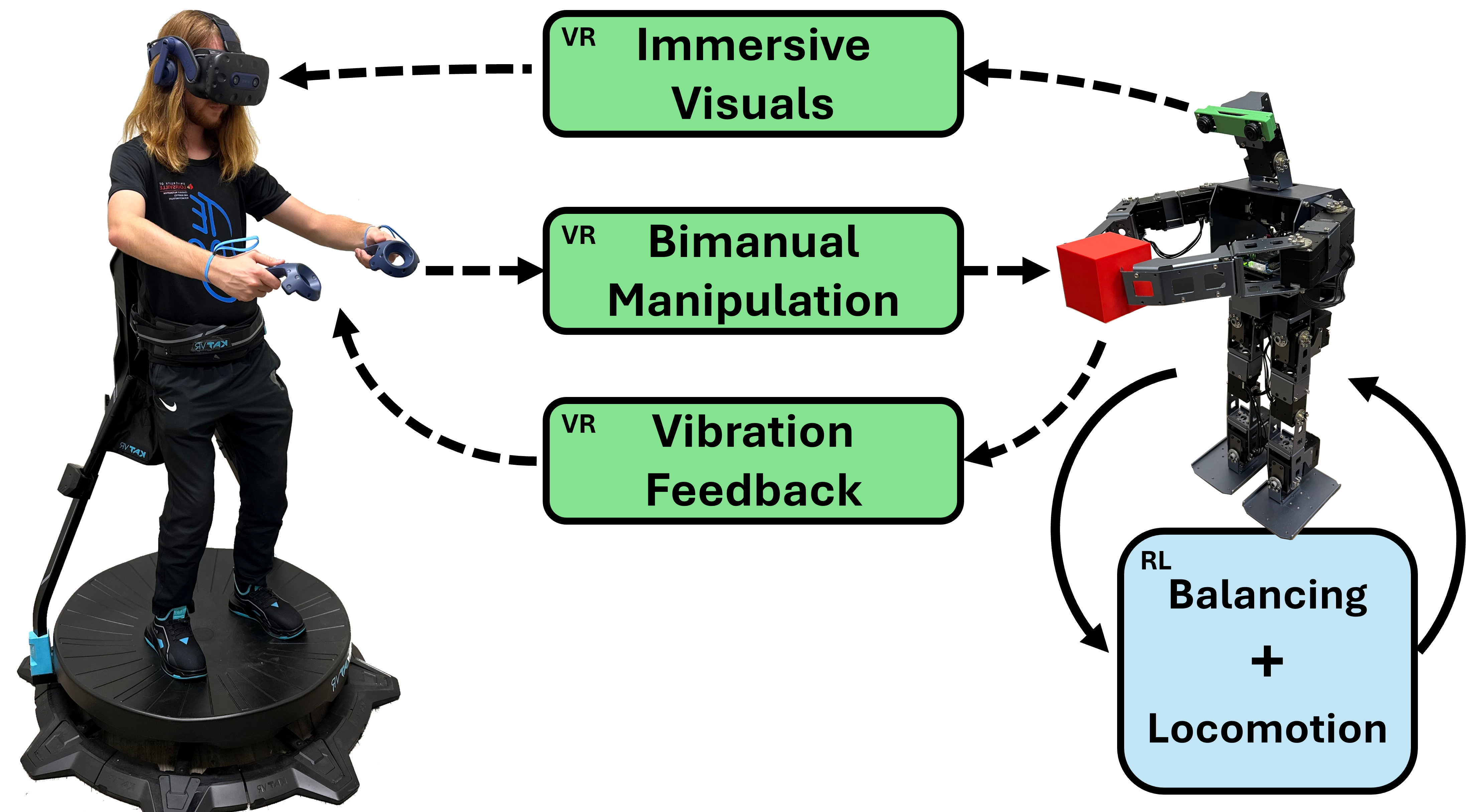}
	\caption{Human teleoperator using the proposed tele-loco-manipulation platform to stack boxes from a distance. Green blocks are associated with VR, while blue relates to RL.}
    \vspace{-4mm}
	\label{fig_Conceptual_Pic}
\end{figure}

In spite of the enormous potential of imitation learning, data remains a prime bottleneck. While many researchers have collaborated for countless hours to develop numerous high-quality teleoperated robot demonstrations \cite{OpenX}, humanoid data is almost never present. Reasons for demonstration disparities include the excessive pricing and difficulty of working with full-scale hardware. Thus, miniature humanoid platforms become an attractive alternative, offering a significantly easier handling process at a fraction of the cost. Hence, smaller platforms are popular within research for prototyping and human-robot social interactions. That said, such hardware commonly possesses fewer manipulator Degrees of Freedom (DoFs) than their full-sized counterparts, harming their capacity for general-purpose manipulation \cite{AvatarDarwin}. As a result, little work has been done on combined locomotion and manipulation---\enquote{loco-manipulation}---of miniature humanoids. Regardless of their abilities for single-armed manipulation, important, relevant, and foundational data for future efforts in loco-manipulation could be obtained from full-body teleoperated manipulation demonstrations of miniature robots. Therefore, a robust and generalizable method for tele-loco-manipulation with small humanoids needs to be created. \textbf{This preliminary work aims to address this need by developing a framework for VR whole-body control of DYNAMIXEL-based humanoids}, beginning with ROBOTIS OP3. Figure \ref{fig_Conceptual_Pic} shows a diagram of the proposed system architecture, leveraging VR and RL for tele-loco-manipulation control.

\subsection*{Paper Contributions}
The developed tele-loco-manipulation system's primary features, alongside this paper's main contributions, are:
\begin{enumerate}
    \item A software architecture for full-body tele-loco-manipulation control of humanoid robots.
    \item Data-driven DYNAMIXEL XM430 actuator modeling for torque control.
    \item A full-body PD-based impedance controller for robot chains made of DYNAMIXEL actuators.
    \item Framework and controller validation via experimentation on the humanoid ROBOTIS OP3.  

\end{enumerate}

\subsection*{Article Structure}
This article is split into six sections: Section II reviews related work in manipulator teleoperation, RL locomotion, and classical loco-manipulation. Section III expounds the system through theory, simulation, robot learning, and manipulation user interfaces. Section IV describes the experimental campaign, while Section V reviews associated data and comments on system efficacy. Lastly, Section VI concludes the work while positing future improvements.

\section{Related Work}

The following subsections cover relevant works in the topics of VR telemanipulation, RL biped control, and loco-manipulation.

\subsubsection{Classical Humanoid Loco-manipulation}
Previous works in classical humanoid loco-manipulation commonly used position-controlled robots that could overpower external system dynamics, allowing physical interactions to be treated as disturbances to the walking gait \cite{handbook, vazPushcart}. However, such approaches commonly hinder whole-body walking stability; as such, \cite{vazWaterbuckets} used explicit dynamic models of objects to update robot gait and manipulator motions to dampen resonant frequencies that resulted in negative outcomes like liquid spillage or falling over. Other common methods demand whole-body controllers for maintenance of Generalized Zero Moment Point (GZMP) \cite{handbook} to ensure the robot cannot fall during a contact-rich interaction.

\subsubsection{VR Robot Telemanipulation}
Enormous developments in VR telemanipulation have been made in the past 5 years. Catalysts such as the COVID-19 pandemic and the ANA Avatar XPRIZE telepresence competition alongside innovations in VR and haptic technology drove contemporary trends for haptic glove-controlled robotic hands and upper-body motion retargeting \cite{XPRIZE, IROS2024robotFROMjapanTOitalyYARP, iCub, HuboXPRIZE, VazJournal}. Such works led to the previous development of a miniature humanoid upper-body geared towards social robotics \cite{AvatarDarwin}. However, these systems all relied on rigidly position-controlled actuation, simplifying operator motion retargeting at the cost of robot safety during contact with the environment or people. In response to this, several works have focused on developing torque-controlled, backdrivable, and compliant agents \cite{G1Hospital, Nadia, openTelevision}. Such capabilities are critical to reliably collecting data for imitation learning policies that can be deployed safely onto real humanoids, as demonstrated by \cite{trill, HOMIE, mobileTelevision}.

\subsubsection{RL Bipedal Locomotion and Sim2Real}
RL is rapidly becoming the dominant method for achieving robust and natural gaits within legged robots. \cite{walkInMins} revealed that massive parallelized simulations run on consumer GPUs could rapidly converge to robust \enquote{blind} walking (locomotion without exteroception) that was deployable zero-shot onto real hardware. In addition to this, RL has seen successful deployment on several small-scale bipedal platforms \cite{op3CompliantRl, tinyRobotRl}. A particularly notable work from this field was from \cite{op3Soccer}, which used adversarial RL to train two robots to play soccer against each other over thousands of years of simulated self-play. This resulted in the robots obtaining gaits that could walk significantly faster than the original optimal-control, ZMP-based, preprogrammed walking routines. Lastly, \cite{ToddlerBot} showcased autonomous loco-manipulation capabilities within a miniature humanoid by independently swapping from diffusion policies for manipulation and RL for locomotion. Interestingly, the policies were trained from data obtained from non-egocentric teleoperation, instead relying on an external human teleoperator that manipulates a gamepad or a leader-follower device. Egocentric demonstrations, bolstered by telepresence control methods, could result in improved generalization for more difficult tasks, as operation from the first-person perspective can endow the operator with additional knowledge of the robot's perceptual shortcomings. 

\label{RelatedWork}

\section{Methodology}
This section details combining independent practices in VR telemanipulation and RL locomotion to realize miniature humanoid whole-body control. Additional implementation information provided regarding the VR user interface, RL training parameters, and \textit{sim2real} techniques for DYNAMIXEL actuators.

\subsection{Embodiment Strategy}
Many VR systems include a Head Mounted Display (HMD) and hand controllers with buttons and 6D pose tracking to keep interactions with virtual objects intuitive. This work follows precedents from \cite{AvatarDarwin, DOLA}, whereby a scaled virtual model of a robot is superimposed over the operator's torso to give an impression of \textit{robotic embodiment}. Furthermore, operator hand positions obtained from the controllers are then fed into an inverse kinematics (IK) solver to generate robot arm motions that mimic operator movements. Consequently, on a humanoid robot, arm motions influence the robot's balance; to counteract this, a robust walking/balancing controller is needed. Such a controller can be realized using RL, particularly by employing random perturbations to the robot during training. Finally, to ensure everything can be done safely for the hardware, compliance can be implemented through joint-side impedance control. Given the target hardware platform, ROBOTIS OP3, this requires hardware modifications and rewriting all robot software (including system-level communications) from scratch.

\subsection{VR Robot Control}
\subsubsection{Hardware}
Telemanipulation control in VR is realized using a VIVE Pro 2 HMD to track operator head movements and VIVE controllers for operator hand poses. Both the HMD and controllers are optically tracked by four SteamVR Basestation 2.0s, as shown in Figure \ref{fig_Conceptual_Pic} (left). Additionally, the operator stands on a KAT Walk C2+ VR treadmill, which enables locomotion retargeting (for a future study). Everything is connected to an i9-based PC with an RTX 4090 GPU to ensure minimal VR motion latency. 

ROBOTIS OP3 possesses twenty revolute DoF (two 6 DoF legs, two 3 DoF arms, one 2 DoF pan/tilt head) actuated directly by ROBOTIS DYNAMIXEL XM430-W350-R servos. Additionally, it houses an i7 NUC and Robotis OpenCR board (with a 9 DoF Inertial Measurement Unit (IMU)) within its chassis. Also, the robot's original camera was replaced with a generic VR180 dual-fisheye lens USB camera to enhance telepresence. Currently, the system relies on a direct connection to wall power.  

\subsubsection{Robot Software}
\textit{Unity} was selected as the chief development software, which provides VR scene creation and physics simulation tools. The virtual environment consists of a model of the robot's upper torso within two spheres that display camera video feeds. State data and commands are transmitted wirelessly between Unity and the robot using ROS \cite{ROS} and the Unity Robotics plugin \cite{unityRobotics}. From there, a multi-objective IK solver \cite{Starke} generates sets of joint positions for each robot arm to track VR controller positions while avoiding joint limits and self-collisions. A benefit of using \cite{Starke} is the inclusion of built-in trapezoidal motion profiles, simplifying enforcement of maximum joint velocities and accelerations. Prior to beginning motion retargeting, the VR robot model is calibrated by scaling the ROBOTIS OP3 model to share the operator's shoulder width and height, as per \cite{AvatarDarwin, DOLA}.

Generated IK solutions are then sent to the robot, which tracks the desired arm positions using a custom PD-based torque controller running at 200 $Hz$ (bandwidth limited by TTL communication speed). Torque control of the DYNAMIXEL XM430 servos is performed open-loop via hardware-level current control. Required currents are calculated by assuming a linear relationship between input current and output shaft torque, wherein the motor constant is dictated by the stall torque/current. In preliminary work, we found highly linear behavior between these two variables on a static testbed. Nevertheless, unlike the legs of the robot, the arms and head use lower PD gains to increase manipulative compliance. Commanded torques are computed below in Equation \eqref{eqn_pd_controller}.

\begin{equation}
	\boldsymbol{\tau}_{cmd} = K_P (\boldsymbol{q}_{des} - \boldsymbol{q}) - K_D \boldsymbol{\dot{q}}
    \label{eqn_pd_controller}
\end{equation}

Here, $\boldsymbol{\tau}_{cmd}$ $\in$ $\mathbb{R}^{20}$ is the vector of command torques sent to the robot's actuators, $\boldsymbol{q}_{des}$ $\in$ $\mathbb{R}^{20}$ are the desired joint positions obtained from Unity, while $\boldsymbol{q}$ $\in$ $\mathbb{R}^{20}$ is joint position feedback. $\boldsymbol{\dot{q}}$ $\in$ $\mathbb{R}^{20}$ are joint velocities obtained from a first-order finite difference of joint positions. Lastly, $K_P$ $\in$ $\mathbb{R}^{20 \times 20}$ and $K_D$ $\in$ $\mathbb{R}^{20 \times 20}$ are diagonal matrices of joint stiffness $[Nm/rad]$ and damping $[Nm/rad-s]$ gains. Contrary to the opaque nature of the position controllers on DYNAMIXEL actuators, this presents a clear physical meaning to actuator parameters, simplifying physics simulation and joint tuning. Moreover, this formulation (alongside the torque controller) is entirely robot-agnostic, permitting application to any machine made of DYNAMIXEL actuators. 

Outside of arm control, neck control is accomplished by observing headset rotations. For instance, pan and tilt rotations of the HMD are sent to the robot's neck mechanism, while headset roll is ignored. Hence, the operator can reposition the robot's cameras by moving their own head. 

\subsubsection{User Interface}
\begin{figure}
    \vspace{2mm}
    \centering
    \includegraphics[width=1.0\linewidth]{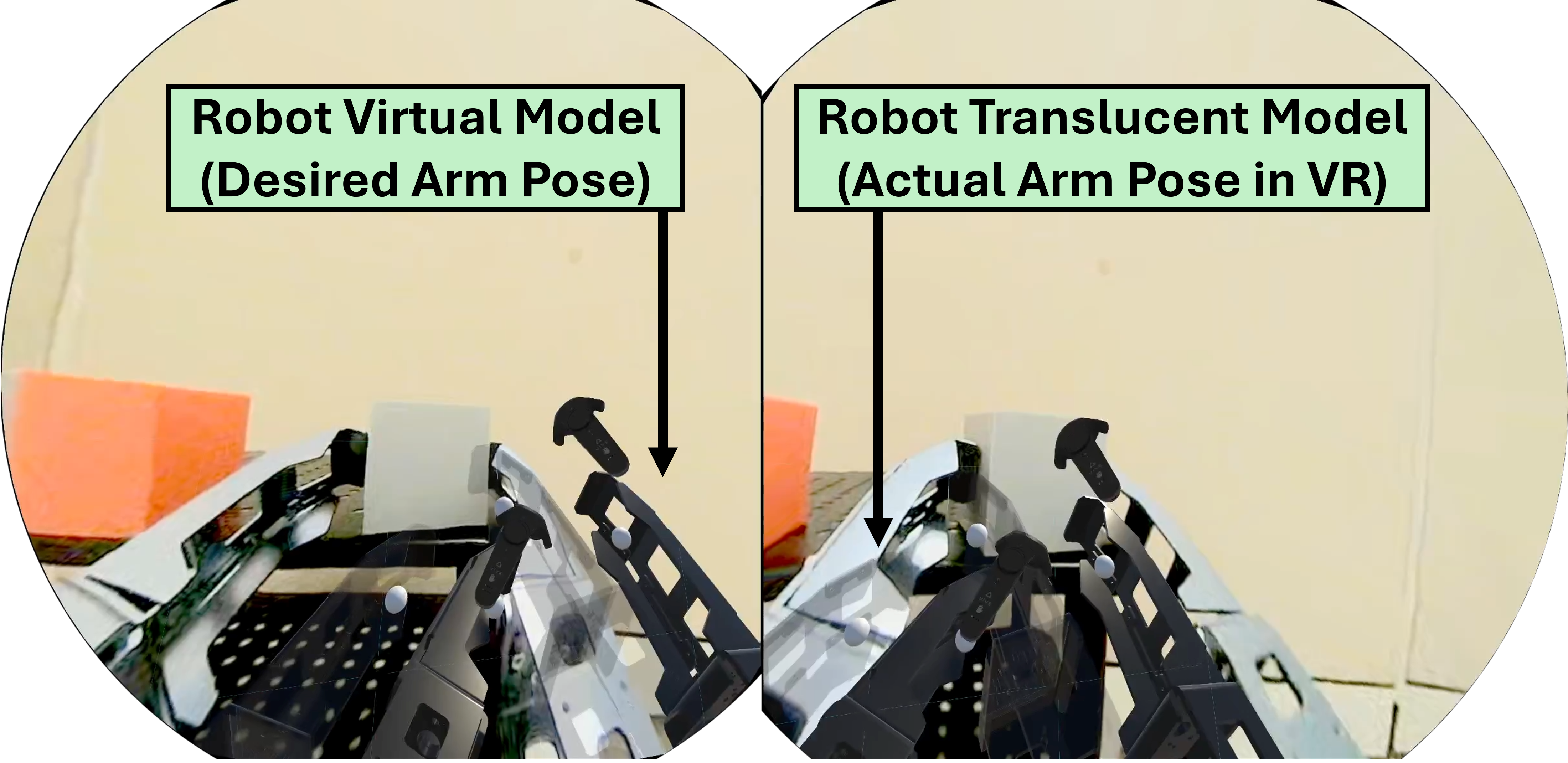}
    \caption{"Operator view" of the remote environment from images streamed directly to the HMD. Additionally, the virtual (and translucent) robot models are shown in dark opaque and transparent gray.}
    \label{fig:UnityView}
    \vspace{-2mm}
\end{figure}

Beyond robot communications and control, a user interface is also critical to telemanipulation. The VR model of the robot's torso tracks the \textit{position} of the headset such that the model and user's shoulders remain approximately coincident (assuming no user torso rotations, as OP3 has no waist DoF). Also, the video stream received from the dual fisheye lens camera (3840 $\times$ 1080 horizontal side-by-side images at 60 $Hz$) is projected within the virtual spheres surrounding the robot, per \cite{Equirectangular, NASAEquirect, NimbroVision, DOLA}, dewarping the barrel-distorted image into a VR180 view that extends visual periphery beyond the field of view (FOV) of the headset, as shown in Figure \ref{fig:UnityView}. The video feed is sent over the User Datagram Protocol (UDP) via a GStreamer pipeline that provides GPU-accelerated H.265 transcoding to permit glass-to-glass latencies $\sim$ 100 $ms$ while using less than a 4 $Mbps$ bitrate. Since a binocular camera is used, operators can naturally perceive depth near the center of the image \cite{AvatarHuboVision}. 

That said, the robot's neck motions have a mechanical latency that imposes a relative delay ($\sim 100$ $ms$) between the operator and robot's head orientation. Such latencies are key contributors to motion sickness in VR. As such, rotational latency is masked by rotating the spheres displaying the robot's egocentric view in the opposite direction. An Exponential Moving Average (EMA) filter is used to smoothen and delay spherical display rotations to better match mechanical delays. 

Arm motion retargeting is toggled on/off using the triggers of the VIVE Controllers. Releasing the triggers drives the maximum acceleration and velocity limits of the IK solver to zero, halting additional movement tracking. Contrarily, pressing the triggers slowly ramps the IK velocities/accelerations back to robot joint limits via an EMA filter. A translucent model of the robot upper torso superimposed over the virtual model the operator manipulates. The translucent model displays robot joint position feedback (received from ROS), enabling the operator to clearly see where the robot is commanded to move versus where it actually is. Rough estimates of contact forces are calculated using Hooke's Law on the distance between the end effectors of the virtual and translucent robots, as described by Equation \eqref{eqn_force_est}:

\begin{equation}
    \boldsymbol{F} =  
    \begin{cases} 
    0 & \text{if } \|\boldsymbol{x}\| < 0 \\
    -K\boldsymbol{x} & \text{if } \|\boldsymbol{x}\| \geq x_{threshold} \\
    \end{cases}
    \label{eqn_force_est}
\end{equation}

Here, $\boldsymbol{F}$ is a 3D vector representing cartesian forces felt by the real robot. $\boldsymbol{x}$ is the vectorial distance between the virtual and translucent robot model end effectors, and $K$ is a static stiffness gain. If the distance between the two models is below a certain distance threshold $x_{threshold}$, the force is set to zero, as it is assumed that small errors are caused by mechanical latency between the commanded and desired joint positions. Unlike Jacobian-based methods, which requires knowledge of full manipulator dynamics to produce reliable results for external forces, this can be applied regardless of how well robot dynamics are known. Procedurally-generated 3D arrows are drawn on the translucent robot's end effector to visualize force expenditure. Lastly, the 2D touchpads on the VIVE Controllers are mapped to differential-drive style velocity commands (one angular and one linear velocity component), which are sent to the robot's lower body for locomotion control.

\subsection{RL Locomotion}

\subsubsection{Formal Definition}
For an RL policy modeled as a Markov Decision Process (MDP), the agent's goal is to maximize a reward function $G_t$, consisting of the current state's $\mathcal{S}$, possible future actions $\mathcal{A}$, the agent's reward $\mathcal{R}$ for a given action in $\mathcal{A}$, and a discount factor $\gamma^k$.   
\begin{equation}
    G_t = \sum^{\infty}_{k=0} \gamma^k \mathcal{R}(\mathcal{S}_{t+k},\mathcal{A}_{t+k}),\label{eq:CDR}
\end{equation}

Where a policy $\pi(\mathcal{A}|\mathcal{S})$ relates the states of the agent to possible actions. Where an optimal policy $\pi^*$ maximizes the functions,
\begin{equation}
    \pi^* \shrteq \arg \max_{\pi}\left[ \mathbb{E}_{\pi} \sum^{\infty}_{k=0} \gamma^k \mathcal{R}(\mathcal{S}_{t},\mathcal{A}_{t}|\mathcal{S}_0 \shrteq \mathcal{S}) \right],
    \label{eq:Vpi}
\end{equation}
\begin{equation}
    \pi^* \shrteq \arg \max_{\pi} \left[ \mathbb{E}_{\pi} \sum^{\infty}_{k=0} \gamma^k \mathcal{R}(\mathcal{S}_{t},\mathcal{A}_{t}|\mathcal{S}_0\shrteq\mathcal{S},\mathcal{A}_0\shrteq\mathcal{A}) \right].
    \label{eq:Gpi}
\end{equation}

Where $\pi^*$ must maximize both the expected reward for a given state $\mathcal{S}$ as seen in Eq.\eqref{eq:Vpi}, and the reward for a given state $\mathcal{S}$ taking an action $\mathcal{A}$ shown in Eq.\eqref{eq:Gpi}.

\subsubsection{Sim2Real and Actuator System Identification}
A significant hurdle in RL is the sim2real gap, often solved by better modeling and domain randomization~\cite{sim2realDR,DowdyCASE2025}. Domain randomization allows for aspects of the robot, such as its mass, to be changed at reset. The locomotion policy produced in this work uses these domain randomization techniques to help cross the sim2real gap. During telemanipulation, the mass and center of mass for the bipedal robot will change, resulting in a new zero-moment point (ZMP). Using domain randomization on link mass, along with the environment's gravity, will produce a more robust policy capable of adapting to mass changes. In simulation, the robot's initial state and foot friction also had their domains randomized along with the velocity of pushing events. These pushing events add a random velocity to the robot's linear velocity, occurring every $(10, 15)$ seconds. The possible values for each of the quantities randomized can be seen in Table \ref{tab:DomRand}.

\begin{table}\centering
\vspace{2mm}
\caption{Domain Randomization}\label{tab:DomRand}
\setlength{\tabcolsep}{4pt}
\renewcommand{\arraystretch}{1.5}
\begin{tabular}{cccl}
\toprule
\textbf{Term}              & \textbf{Operation} & \textbf{Range}\\ 
\midrule
Foot Friction  &  New    & $(0.5, 0.9)$
\\
Gravity    & Scale   & $(0.95,1.3)$                    \\
Link Mass  & Scale        & $(0.7,1.3)$         \\
Body Velocity & Add & $(\text{-}\hspace{0.1em}0.25, 0.25)$ \\
Body Position & Add & $(\text{-}\hspace{0.1em}0.5,0.5)$ \\
Body Orientation & Add & $(\text{-}\hspace{0.1em}0.02,0.02)$ \\
Joint Positions  & Scale  & $(\text{-}\hspace{0.1em}0.3, 0.3)$ \\
Joint Velocities  & Scale  & $(\text{-}\hspace{0.1em}2.5, 2.5)$ \\
\bottomrule
\end{tabular}
\vspace{-2mm}
\end{table}

\begin{figure}
    \centering
    \includegraphics[width=0.9\linewidth]{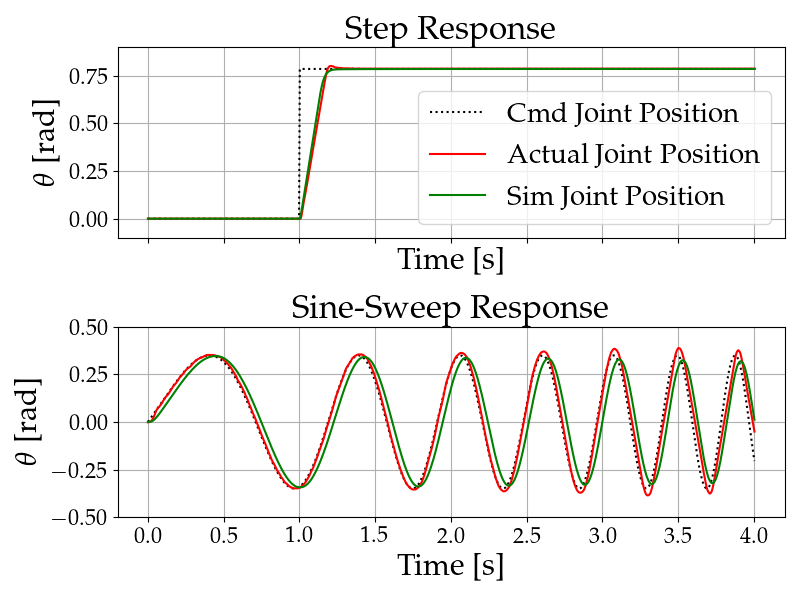}
    \caption{Real2Sim PD gain tuning for the robot's actuators (Dynamixel XM430-W350). The plots compare real and simulated actuator step and sine-sweep trajectory responses.}
    \label{fig:PD2sim}
    \vspace{-2mm}
\end{figure}

An accurate model of actuator dynamics was crucial to achieve a robust locomotion policy with a zero-shot transfer. Both Isaac Sim and Isaac Lab have actuator models, with Isaac Lab allowing for more detail in actuator dynamics. Initial manual tuning of PD parameters in simulation was done, using sinusoidal and step input responses collected from the Dynamixel with the developed torque controller. This manual tuning resulted in a locomotion policy transfer; however, minor instabilities at high joint velocities were seen. This led to the creation of a pendulum testing bench to model the dynamics of the actuator at various point masses and PD gains. To accurately estimate the joint parameters, the \textit{Better Actuator Modeling} (BAM) \cite{BAM} method was used. Parameter estimation based on the pendulum test bench data produced an armature (0.045), friction loss (0.03), and effort limit (3.6). These values were then verified in Isaac Sim, shown in Figure \ref{fig:PD2sim}.

\subsubsection{Observations and Rewards}
Through observation and reward shaping, the RL policy becomes robust enough to counteract the forces and torque applied to the robot during teleportation. This section covers the considerations when designing the observation and rewards for this bipedal locomotion policy. 

The observation space comprises five terms, with three of these using proprioceptive data for angular velocity, projected gravity in body frame, and joint positions, with a detailed description shown in Table \ref{tab:obsTerms}. Each observation term is stacked with a previous nine-timestep history, represented as a $330\times1$ vector. Stacking previous observed states proved to be more successful during the sim2real transfer, producing less noise in actions during inference. Simultaneously, joint velocities can be inferred by the model, removing it from its observation space.

\begin{figure*}[b]
	\centering
	\vspace{-2mm}
	\includegraphics[width=0.45 \linewidth]{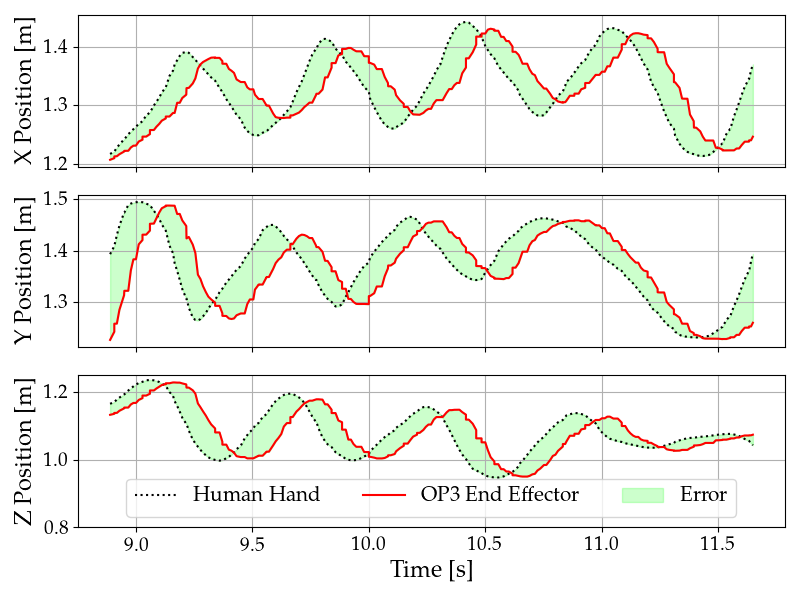}
	\includegraphics[width=0.45 \linewidth]{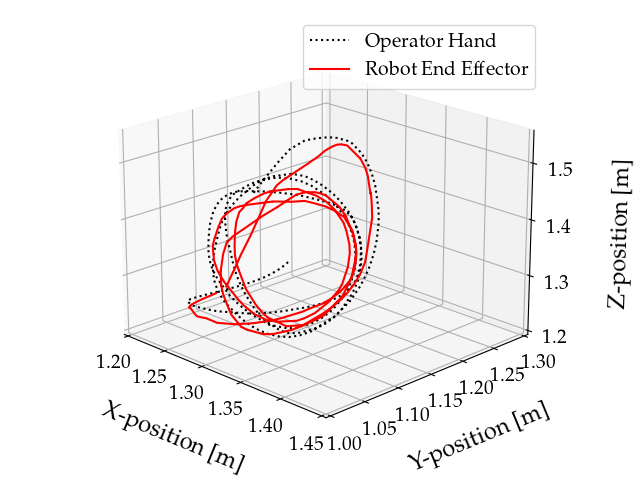}
    \vspace{-2mm}
	\caption{Results from telemanipulation experiments, comparing human hand to robot end effector motion. The plots on the left show transient data in the $x$, $y$, and $z$ directions, whereas the plot on the right shows the resulting circular 3D trajectories.}
	\label{fig:fig_arm_transient_plots}
    \vspace{-2mm}

\end{figure*}

To achieve stable locomotion during the robot's teleoperation, several custom rewards were created outside of the standard and pre-baked into Isaac-Lab. As the center of mass during teleoperation changes significantly, the ability for this policy to adjust its feet quickly is desirable. Trying to establish this relationship expressed as a significant penalty on projected gravity $xy$ components in the foot frame. The intuition behind this is as follows: when the center of mass changes, the orientation of the body is likely to change; however, if only a minor change occurs, the policy is likely to stay in this state due to a cost of moving, with the foot orientation penalty aiming to shift this and allow the network to make these adjustments as needed. A custom gait reward term was also created, which rewards the agent for the dual support and single support phases. Table \ref{tab:rwdTerms} shows more detail on each reward, with a general expression and assigned weight. This reward shaping guides the policy towards dynamic walking, taking advantage of the gained momentum to minimize several of its energy minimization terms.

\begin{table}\centering
\caption{Observation Space Terms}\label{tab:obsTerms}
\setlength{\tabcolsep}{4pt}
\renewcommand{\arraystretch}{1.5}
\begin{tabular}{cccl}
\toprule
\textbf{Observation Term}               & \textbf{State Expression}         & \textbf{Injected Noise}    \\
\midrule
Angular Velocity        & $\boldsymbol{\omega}_{yaw}$ $\in \mathbb{R}^3$ & 0.2 
\\

Projected Gravity       & $\textbf{G}_{base} \in \mathbb{R}^3$    &0.1                      \\
User Command          &$ \textbf{u}^*_{base}\in\mathbb{R}^3$    &0.0                       \\
Joint Positions         & \textbf{$\left|\left|\theta_i-\theta\right|\right|$} $\in \mathbb{R}^{12}$   &0.05                      \\
Previous Action      & \textbf{$\mathcal{A}_{t-1.}$} $\in \mathbb{R}^{12}$      &0.0                       \\
\bottomrule
\end{tabular}    
\end{table}

\begin{table}\centering
\vspace{2mm}
\caption{Rewards and Penalties}\label{tab:rwdTerms}
\setlength{\tabcolsep}{4pt}
\renewcommand{\arraystretch}{1.7}
\begin{tabular}{cccl}
\toprule
\textbf{Reward Term}                    & \textbf{General Expression}        & \textbf{Weight}       \\
\midrule
Linear Velocity Error        & $\mathit{exp}(-(\left|\left|\upsilon_{x,y}^{\mathit{des}} - \upsilon_{x,y}^{\mathit{cur}}\right|\right| )$      &2.0          \\
Angular Velocity Error  &$\mathit{exp}(-(\left|\left|\omega_{\mathit{z}}^{\mathit{des}} - \omega_{\mathit{z}}^{\mathit{cur}}\right|\right|/2) )$         &1.0 \\
Gait                         & $\prod^2_{i=1}f_{\mathit{sync},i}\cdot\prod^4_{i=1}f_{\mathit{async},i}$                     &0.75  \\
Foot Clearance                & $\mathit{exp}(-\sum^4_{i=1}(f_{\mathit{z},i} - 0.04))$                              &1.0           \\
\midrule
\textbf{Penalty Term}                   & \textbf{General Expression}                                           & \textbf{Weight}       \\
\midrule
Action Smoothness  & $||\mathcal{A}_t - \mathcal{A}_{t\text{-}1}||$            &-0.15         \\
Action Rate  & $||\mathcal{A}_t||$            &-0.125         \\
Base Motion                      & $| \omega_x + \omega_y |$      &-2.0         \\
Base Orientation                 & $||\textbf{G}_{x,y}||$                           &-5.0           \\
Foot Orientation & $||\textbf{G}_{x,y}||$  &-5.0           \\
Foot Slippage                    & $||\textbf{C}_{\mathit{foot}} \cdot \upsilon_{\mathit{foot}}||$                         &-2.0           \\
Joint Torque                     & $||\tau||$                          &-0.002           \\
Joint Velocity                   & $||\dot{\theta}||$                           &-0.002           \\
Joint Acceleration          & $||\ddot{\theta}||$                           & -0.000025
\\
Joint Deviation             & $\sum^{12}_{i=1}(\theta_{\mathit{i}}-\theta)$  &-0.125 \\
Joint Position Limit       & $\sum^{12}_{h=1}(\theta\geq\theta_{\mathit{max}})$  &-10.0 \\
Termination Penalty        & $\textbf{C}_{\mathit{Arm, Body, Head}}$ &-10.0
\\

\bottomrule
\end{tabular}
\vspace{-2mm}
\end{table}

\section{Experimentation}
Validation of subsystems alongside a demonstration of the full system in operation is shown within this section. A sample trial of a bimanual tele-loco-manipulation experiment can be viewed at: \break \url{https://www.youtube.com/watch?v=FI2gyPqU5xI}

\subsection{Telemanipulation Validation}
To gauge telemanipulation performance (specifically in regards to operator motion tracking latency and accuracy), a simple motion experiment is used. This experiment is comprised of the operator rapidly tracing a circle with both arms ten times. The positions of the VR controllers and the robot's actual end effectors are both recorded at 50 $Hz$. 

\subsection{Locomotion Validation}
Evaluation of the trained locomotion policy in terms of performance and stability is accomplished by sending random whole-body velocity commands to the robot every three seconds for a total of eighty seconds.  Normalized projected gravity is a metric of bipedal stability, as it symbolizes the posture of the humanoid during locomotion. As such, the normalized projected gravity is recorded alongside the commanded velocities during this experiment.

\subsection{Tele-Loco-Manipulation Experiment}
Lastly, to assess the full system's capacity for tele-loco-manipulation, a block relocation experiment requiring simultaneous bimanual manipulation and locomotion is devised. The experiment requires a teleoperated humanoid to traverse a short distance ($<1$ $m$), pick up a 3D-printed PLA cube (weighing approximately 40 $g$) with both arms, walk to a specified container, drop the cube into it, and repeat the process. Unlike the prior experiments, which aimed to analyze specific quantitative performance metrics, this scrutinizes how well the full-stack works together. That being said, numeric data regarding task completion time, walking speed, and number of blocks relocated successfully are recorded. Ten trials are completed in total.

\begin{figure*}[t!]
	\centering
	\vspace{2mm}
	\includegraphics[width=\linewidth]{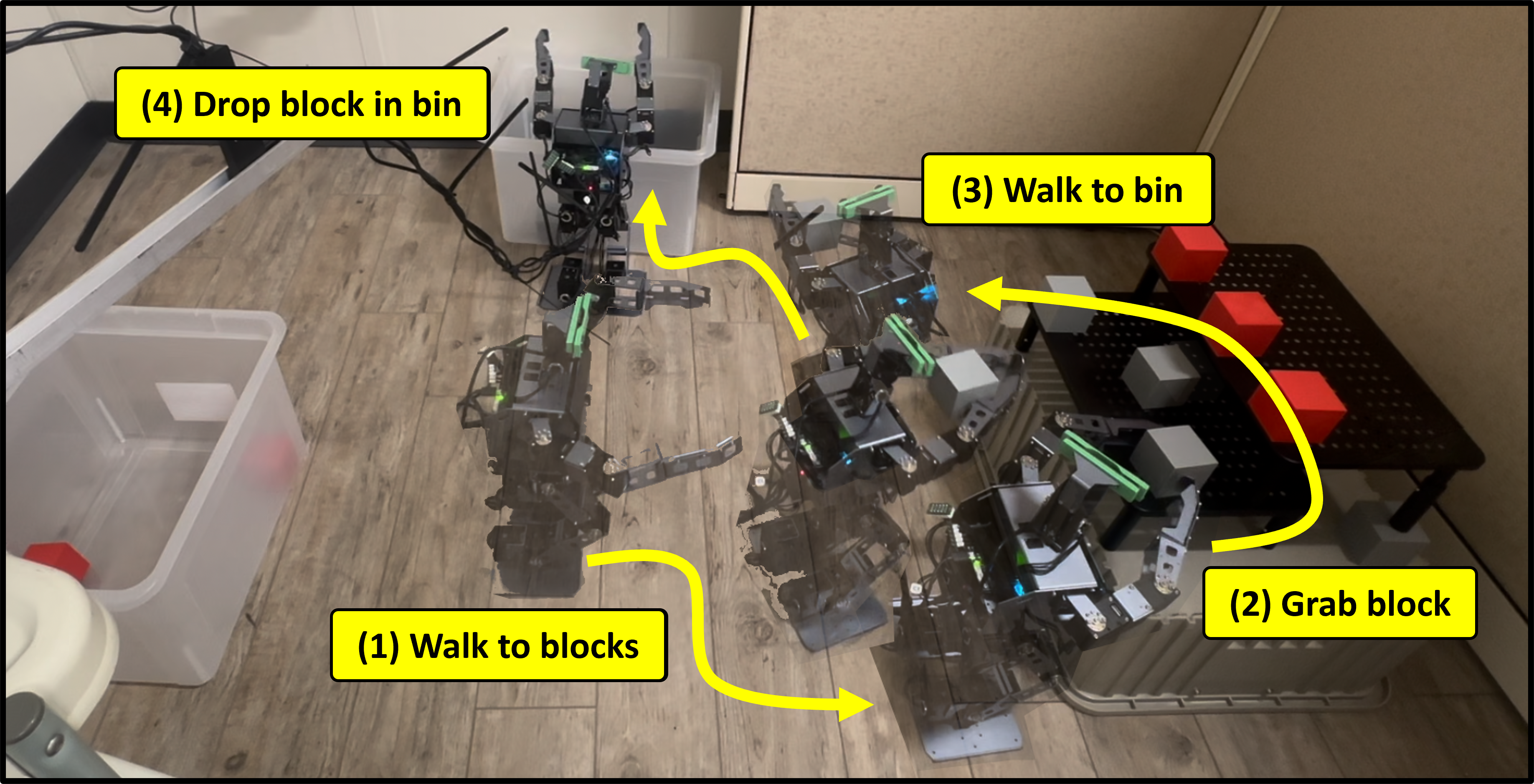}
	\caption{Snapshots from a real tele-loco-manipulation experiment trial that visually depict the act of block relocation.}
    \vspace{-4mm}
	\label{fig:BigPic}
\end{figure*}

\begin{figure}[b!]
    \centering
    \vspace{-2mm}
    \includegraphics[width=\linewidth]{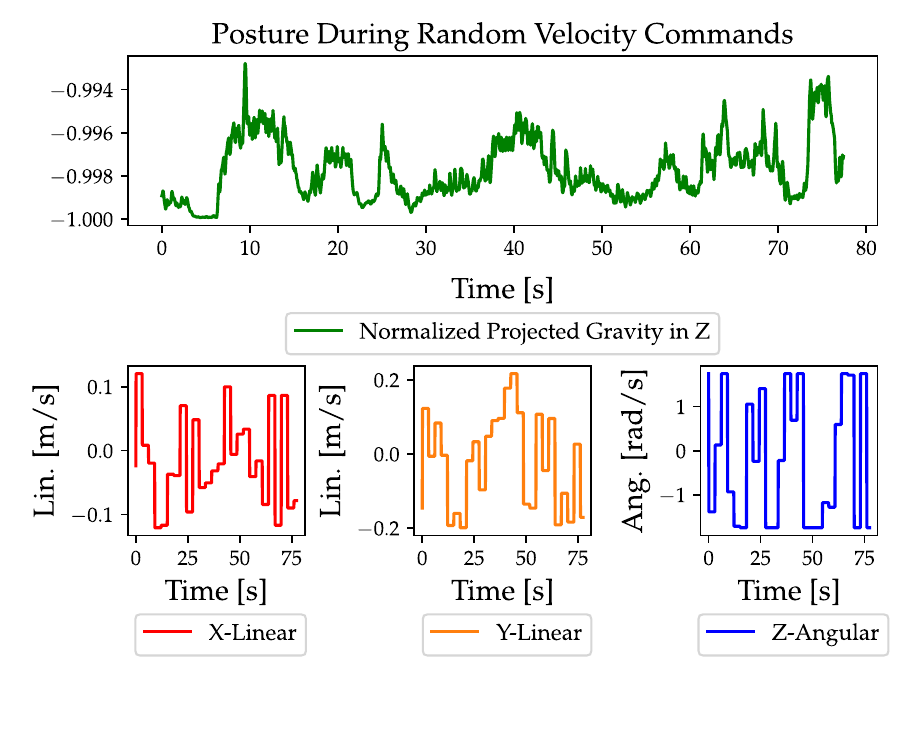}
    \caption{Humanoid robot's projected gravity during random velocity command testing.}
    \label{fig:CMDCsGrav}
\end{figure}

\section{Results and Discussion}
This section analyzes and displays data collected from the aforementioned experiments. Following this, the implications of the results are discussed, with system limitations being pointed out explicitly.

\subsection{Experimental Results}

\subsubsection{Telemanipulation Validation}
The performance of a telemanipulation system primarily stems from motion latency and accuracy. End-to-end motion latency is found via calculation of the cross-correlation between transient human and robot hand positions. The motion latency is defined as the peak of the cross-correlation function for the two waveforms. When applying this analysis to the $y$-direction, a latency of 220 $ms$ is found. This agrees with values found in Figure \ref{fig:fig_arm_transient_plots} (left), which shows a horizontal displacement of about 200 $ms$ between the human and robot motions in any direction. Notwithstanding latency, the circular trajectories traced by the system are visualized in Figure \ref{fig:fig_arm_transient_plots} (right), which qualitatively demonstrate that the humanoid tracks operator motion. Quantitatively speaking, the Root Mean Squared Error (RMSE) and Mean Absolute Error (MAE) between the normalized positions of the operator's/robot's right arms are listed in Table \ref{table_Tracking}. 

\begin{table}[t]
    \vspace{2mm}
	\renewcommand{\arraystretch}{1.39}
	\caption{\emph{Normalized Arm Position Tracking Error}}
	\centering
	\begin{tabular}{|c| c| c| c| }
		\hline
        Hand & Dimension & MAE [$\%$] & RMSE [$\%$]\\
        \hline
        Right & $x$ & 8.76 & 12.35 \\
        \hline
        Right & $y$ & 10.58 & 15.07 \\
        \hline
        Right & $z$ & 8.68 & 11.17 \\
        \hline
    \end{tabular}
    \label{table_Tracking}
    \vspace{-2mm}
\end{table}

\subsubsection{Locomotion Validation}
The robustness of locomotion policies is tested in simulation by applying random body velocity perturbations to show the system's capability to reject these disturbances and continue standing. This is represented as a fast change in the center of mass in our RL locomotion performance measurement, where the projected gravity or posture of the robot stays close to its normalized value. The resulting data from this experiment is presented in the plots of Figure \ref{fig:CMDCsGrav}, where it can be seen that the locomotion policy maintains an upright posture for OP3, even during significant changes in commanded body velocities.  

\subsubsection{Tele-Loco-Manipulation Experiment}
The final tele-loco-manipulation experiment is shown graphically in Figure \ref{fig:BigPic}. Within this, the teleoperated robot was able to correctly relocate 2 out of 6 blocks within 10 $mins$ of trial time. This demanded locomoting a distance of about 5 $m$ per trial, with an average walking velocity around 0.35 $m/s$. 

\subsection{Discussion}
Despite some of the arm positioning errors from the first experiment reaching over 15 $\%$ RMSE or MAE, much of the instantaneous error can be attributed to the motion latency ($\sim 220$ $ms$). Nonetheless, the upper body is clearly capable of reproducing the circular arm motions demanded by the operator. As for locomotion, the normalized projected gravity remains comfortably near the desired value of -1, as desired. This is a solid indication that the RL policy is properly maintaining a stable torso orientation regardless of the abrupt and violent shifting walking velocity commands. Finally, the system struggled to relocate more than two blocks within 10 $mins$ of operation. This slow performance is attributed to a mixture between operational difficulty/frustration. Key sources of operator frustration were related to: the RL-basd locomotion policy not lifting its feet high enough during walking; users needing to cross their arms to generate enough force to lift cubes; locomotion challenges due to the robot's tethered power connection.

\section{Conclusion and Future Work}
This work presents the foundations of a tele-loco-manipulation system for DYNAMIXEL-based miniature humanoid robots. Leveraging VR telepresence manipulation with RL locomotion/balancing alongside a custom-built torque controller enables the upper and lower body of the robot to be controlled independently, simplifying development. The trained RL policy uses heavy domain randomization, base perturbations, and randomized arm movement to improve balancing robustness during object manipulation. Subsystems are independently validated and measured for performance, demonstrating arm motion latencies of $220$ ms. Moreover, the end-to-end system is evaluated with a tele-loco-manipulation block collection experiment. Despite the system struggling during the tele-loco-manipulation experiment (only managing to successfully relocate an average of 2 out of 6 blocks in 10 $min$), the framework still shows potential for whole-body robot control. Additionally, the methods presented can be applied to different robot morphologies made of DYNAMIXEL actuator chains. 

All this being said, significant room for improvement remains. For instance, humanoid proprioceptive walking can be greatly improved by using more difficult terrain, additional reward shaping, etc. Another massive challenge within the work is related to sim2real transfer; the currently-employed actuator model is rudimentary, which demands excessive domain randomization for the RL policy to deal with. Notwithstanding locomotion, the telemanipulation system was too \textit{soft} during trials, requiring the operator to cross their arms during the final experiment to produce enough force to hold onto the blocks. This could be remedied using actual grippers or sensors on the end effectors to ensure adequate contact force to prevent slippage. Additionally, the lack of gravity and friction compensation terms required higher PD gains to be used, ultimately diminishing system compliance. Hence, adding gravitational and friction compensation torques to Equation \eqref{eqn_pd_controller} can improve joint impedance, provided an accurate dynamic model of the robot and its actuators. Beyond this, we plan to integrate the VR treadmill, "KAT WALK", introducing a much more intuitive and immersive experience for tele-loco-manipulation tasks. Further improvements to the vision motion compensation system and the ability for tetherless operation of OP3, remedy several of the pain points of the system, increasing task efficiency. By addressing these challenges, similar systems can become significantly more robust and user-friendly, empowering miniature humanoid tele-loco-manipulation systems to tackle a broad suite of tasks in the future. 

\label{Conclusion}


\begin{thebibliography}{99}

\bibitem{1X}
1X Technologies, \enquote{Introducing NEO Gamma,} \textit{1X Technologies}, Feb. 21, 2025. [Online]. Available: https://www.1x.tech/discover/introducing-neo-gamma

\bibitem{Unitree} 
Unitree Robotics, "Unitree G1 Humanoid", \textit{Unitree Robotics}. [Online]. Available: https://www.unitree.com/g1/

\bibitem{nuclearSuperPaper}
Z. Gu \textit{et al.}, \enquote{Humanoid Locomotion and Manipulation: Current Progress and Challenges in Control, Planning, and Learning}, 2025, arXiv:2501.02116.

\bibitem{OpenX}
A. O'Neill \textit{et al.}, \enquote{Open X-Embodiment: Robotic Learning Datasets and RT-X Models : Open X-Embodiment Collaboration,} \textit{2024 IEEE International Conference on Robotics and Automation (ICRA)}, Yokohama, Japan, 2024, pp. 6892-6903, doi: 10.1109/ICRA57147.2024.10611477.

\bibitem{AvatarDarwin}
A. Dave, J. C. Vaz, J. Kim, N. Kosanovic, N. Kassai and P. Y. Oh, \enquote{Avatar-Darwin a Social Humanoid with Telepresence Abilities Aimed at Embodied Avatar Systems,} \textit{2022 IEEE-RAS 21st International Conference on Humanoid Robots (Humanoids)}, Ginowan, Japan, 2022, pp. 47-52, doi: 10.1109/Humanoids53995.2022.10000176.

\bibitem{handbook}
B. Sciliano and O. Khatib, \emph{Springer Handbook of Robotics}, 1st ed., Heidelberg, Springer Berlin, 2008, pp. 719-739. [Online]. Available: https://doi.org/10.1007/978-3-540-30301-5 

\bibitem{vazPushcart}
J. C. Vaz and P. Y. Oh, \enquote{Expanding Humanoid's Material-Handling Capabilities using Capture Point Walking,} \textit{2020 American Control Conference (ACC)}, Denver, CO, USA, 2020, pp. 2082-2087, doi: 10.23919/ACC45564.2020.9147541.

\bibitem{vazWaterbuckets}
J. C. Vaz and P. Oh, \enquote{Model-Based Suppression Control for Liquid Vessels Carried by a Humanoid Robot While Stair-Climbing,} \textit{2020 IEEE 16th International Conference on Automation Science and Engineering (CASE)}, Hong Kong, China, 2020, pp. 1540-1545, doi: 10.1109/CASE48305.2020.9216826.

\bibitem{XPRIZE}
K. Hauser \emph{et al.}, \enquote{Analysis and Perspectives on the ANA Avatar XPRIZE Competition}, in \emph{Intl. Jour. of Soc. Robotics}, Jan. 2024. Accessed: Feb. 19, 2025. doi: 10.48550/arXiv.2401.05290 [Online].

\bibitem{IROS2024robotFROMjapanTOitalyYARP}
M. Elobaid \textit{et al.}, \enquote{Remote telepresence over large distances via robot avatars: case studies,} \textit{2024 IEEE Conference on Telepresence}, Pasadena, CA, USA, 2024, pp. 183-187, doi: 10.1109/Telepresence63209.2024.10841750.

\bibitem{iCub}
S. Dafarra \textit{et al.}, \enquote{iCub3 avatar system: Enabling remote fully immersive embodiment of humanoid robots,} in \textit{Sci. Robot.}, vol. 9, eadh3834, 2024, doi: 10.1126/scirobotics.adh3834.

\bibitem{HuboXPRIZE}
B. Kim, N. Kassai, Z. Castrejon, N. Kosanovic, J. C. Vaz and P. Oh, \enquote{Approach of Team Avatar-Hubo to the ANA Avatar XPRIZE Finals,} \textit{2024 33rd IEEE International Conference on Robot and Human Interactive Communication (ROMAN)}, Pasadena, CA, USA, 2024, pp. 37-42, doi: 10.1109/RO-MAN60168.2024.10731240.

\bibitem{VazJournal}
J. Vaz, N. Kosanovic and P. Oh, \enquote{ART: Avatar Robotics Telepresence---the future of humanoid material handling loco-manipulation,} in \emph{Intel Serv Robotics}, Dec. 2023. Accessed: Feb. 2025. doi: 10.1007/s11370-023-00499-x [Online]. 

\bibitem{G1Hospital}
S. Atar \textit{et al.}, \enquote{Humanoids in Hospitals: A Technical Study of Humanoid Surrogates for Dexterous Medical Interventions}, 2025, arXiv:2503.12725.

\bibitem{Nadia}
S. Bertrand \textit{et al}., \enquote{High-Speed and Impact Resilient Teleoperation of Humanoid Robots,} Sept. 2024, arXiv:2409.04639 [cs.RO].

\bibitem{openTelevision}
X. Cheng, J. Li, S. Yang, G. Yang and X. Wang, \enquote{Open-TeleVision: Teleoperation with Immersive Active Visual Feedback}, 2024, arXiv:2407.01512.

\bibitem{trill}
M. Seo \textit{et al.}, \enquote{Deep Imitation Learning for Humanoid Loco-manipulation Through Human Teleoperation,} \textit{2023 IEEE-RAS 22nd International Conference on Humanoid Robots (Humanoids)}, Austin, TX, USA, 2023, pp. 1-8, doi: 10.1109/Humanoids57100.2023.10375203.

\bibitem{HOMIE}
Q. Ben \textit{et al.}, \enquote{HOMIE Humanoid Loco-Manipulation with Isomorphic Exoskeleton Cockpit}, 2025 Robotics: Science and Systems, 2025, arXiv: arXiv:2502.13013.

\bibitem{mobileTelevision}
C. Lu \textit{et al.}, \enquote{Mobile-TeleVision: Predictive Motion Priors for Humanoid Whole-Body Control}, 2025, arXiv:2412.07773.

\bibitem{walkInMins}
N. Rudin, D. Hoeller, P. Reist and M. Hutter, \enquote{\textit{Learning to Walk in Minutes Using Massively Parallel Deep Reinforcement Learning}}, \textit{2021 Conference on Robot Learning (CoRL)}, 2021, doi: arXiv:2109.11978.  

\bibitem{op3CompliantRl}
S. Masuda and K. Takahashi, \enquote{Sim-to-Real Transfer of Compliant Bipedal Locomotion on Torque Sensor-Less Gear-Driven Humanoid}, 2025, arXiv:2204.03897.

\bibitem{tinyRobotRl}
S. Katayama, Y. Koda, N. Nagatsuka and M. Kinoshita \enquote{Learning Bipedal Locomotion on Gear-Driven Humanoid Robot Using Foot-Mounted IMUs}, 2025, arXiv:2504.00614.

\bibitem{op3Soccer}
T. Haarnoja \textit{et al.}, \enquote{Learning agile soccer skills for a bipedal robot with deep reinforcement learning}, \textit{Science Robotics}, vol. 9, Apr. 2024, doi: 10.1126/scirobotics.adi8022. 

\bibitem{ToddlerBot}
H. Shi, W. Wang, S. Song and C. K. Liu, \enquote{ToddlerBot: Open-Source ML-Compatible Humanoid Platform for Loco-Manipulation}, 2025 \textit{in arXiv}, doi: 10.48550/arXiv.2502.00893.

\bibitem{DOLA}
N. Kosanovic and J. Vaz, \enquote{A Virtual Reality Framework for Safe Global Bimanual Telepresence
}, \textit{2025 IEEE 21st International Conference on Automation Science and Engineering (CASE)}, Los Angeles, CA, USA, 2025.

\bibitem{ROS}
M. Quigley \emph{et al.}, \enquote{ROS: an open-source Robot Operating System,} \emph{2009 ICRA Workshop on Open Source Software}, 2009. 

\bibitem{unityRobotics}
\enquote{Unity-Robotics-Hub,} GitHub repository, 2022. [Online]. Available: https://github.com/Unity-Technologies/Unity-Robotics-Hub.

\bibitem{Starke}
S. Starke, N. Hendrich and J. Zhang, \enquote{Memetic Evolution for Generic Full-Body Inverse Kinematics in Robotics and Animation,} \textit{in IEEE Transactions on Evolutionary Computation}, vol. 23, no. 3, pp. 406-420, June 2019, doi: 10.1109/TEVC.2018.2867601. 

\bibitem{Equirectangular}
P. Bourke, \enquote{Converting a fisheye image into a panoramic, spherical
or perspective projection,} 2016. Accessed: Feb. 19, 2025. [Online].
Available: https://paulbourke.net/dome/fish2/

\bibitem{NASAEquirect}
S. -H. Berndt et al., \enquote{From Universe to Metaverse: A Leap Into Virtual Collaboration at NASA JPL,} in \textit{IEEE Transactions on Industrial Cyber-Physical Systems}, vol. 1, pp. 287-306, 2023, doi: 10.1109/TICPS.2023.3327948.

\bibitem{NimbroVision}
M. Schwarz and S. Behnke, \enquote{Low-Latency Immersive 6D Televisualization with Spherical Rendering,} \textit{2020 IEEE-RAS 20th International Conference on Humanoid Robots (Humanoids)}, Munich, Germany, 2021, pp. 320-325, doi: 10.1109/HUMANOIDS47582.2021.9555797.

\bibitem{AvatarHuboVision}
J. C. Vaz, A. Dave, N. Kassai, N. Kosanovic and P. Y. Oh, "Immersive Auditory-Visual Real-Time Avatar System of ANA Avatar XPRIZE Finalist Avatar-Hubo," \textit{2022 IEEE International Conference on Advanced Robotics and Its Social Impacts (ARSO)}, Long Beach, CA, USA, 2022, pp. 1-6, doi: 10.1109/ARSO54254.2022.9802964.

\bibitem{sim2realDR}
J. Tan, T. Zhang, E. Coumans, A. Iscen, Y. Bai, D. Hafner, S. Bohez, and V. Vanhoucke, \enquote{Sim-to-Real: Learning Agile Locomotion For Quadruped Robots}, 2018, arXiv:1804.10332

\bibitem{DowdyCASE2025}
J. Dowdy and J. Chagas Vaz, \enquote{Isaac Sim-to-Real: Reinforcement Learning based Locomotion for Quadrupeds,} in \textit{2025 IEEE 21st International Conference on Automation Science and Engineering (CASE)}, Los Angeles, CA, USA, 2025, pp. 2194--2199, doi: 10.1109/CASE58245.2025.11163761.

\bibitem{BAM}
M. Duclusaud, G. Passault, V. Padois and O. Ly, \enquote{Extended Friction Models for the Physics Simulation of Servo Actuators}, 2025, arXiv:2410.08650.

\end{thebibliography}
\end{document}